\documentclass[conference]{IEEEtran}
\IEEEoverridecommandlockouts
\usepackage{cite}
\usepackage{booktabs}
\usepackage{float} 
\usepackage{amsmath,amssymb,amsfonts}
\usepackage{algorithmic}
\usepackage{graphicx}
\usepackage{textcomp}
\usepackage{xcolor}
\def\BibTeX{{\rm B\kern-.05em{\sc i\kern-.025em b}\kern-.08em
    T\kern-.1667em\lower.7ex\hbox{E}\kern-.125emX}}

\makeatletter
\newcommand{\linebreakand}{%
  \end{@IEEEauthorhalign}
  \hfill\mbox{}\par
  \mbox{}\hfill\begin{@IEEEauthorhalign}
}
\makeatother

\begin{document}

\title{Advanced User Credit Risk Prediction Model using LightGBM, XGBoost and Tabnet with SMOTEENN\\

}

\author{
\IEEEauthorblockN{Chang Yu$^*$}
\IEEEauthorblockA{
    \textit{Northeastern University} \\
    Boston, MA, 02115, USA\\
    chang.yu@northeastern.edu}
\and
\IEEEauthorblockN{Yixin Jin}
\IEEEauthorblockA{
    \textit{University of Michigan, Ann Arbor}\\
    Ann Arbor, MI 48109, USA \\
    jyx0621@gmail.com}
\and
\IEEEauthorblockN{Qianwen Xing}
\IEEEauthorblockA{
    \textit{University of Chicago} \\
    Chicago, IL, 60637, USA                \\
    xqw3669@gmail.com}
\and
\IEEEauthorblockN{Ye Zhang}
\IEEEauthorblockA{
    \textit{University of Pittsburgh} \\
    Pittsburgh, PA, 15203, USA\\
    yez12@pitt.edu}
\linebreakand 
\IEEEauthorblockN{Shaobo Guo}
\IEEEauthorblockA{
    \textit{Illinois Institute of Technology}\\
    Chicago, IL 60616, USA \\
    seanguo2017@gmail.com}
\and
\IEEEauthorblockN{Shuchen Meng}
\IEEEauthorblockA{
    \textit{Central University of Finance and Economics} \\
    Beijing, China \\
    scmeng19@163.com}
}
\maketitle

\begin{abstract}
Bank credit risk is a significant challenge in modern financial transactions, and the ability to identify qualified credit card holders among a large number of applicants is crucial for the profitability of a bank's credit card business. In the past, screening applicants' conditions often required a significant amount of manual labor, which was time-consuming and labor-intensive. Although the accuracy and reliability of previously used ML models have been continuously improving, the pursuit of more reliable and powerful AI intelligent models is undoubtedly the unremitting pursuit by major banks in the financial industry. In this study, we used a dataset of over 40,000 records provided by a commercial bank as the research object. We compared various dimension reduction techniques such as PCA and T-SNE for processing high-dimensional datasets and performed in-depth adaptation and tuning of distributed models such as LightGBM and XGBoost, as well as deep models like Tabnet. After a series of research and processing, we obtained excellent research results by combining SMOTEENN with these techniques. The experiments demonstrated that LightGBM combined with PCA and SMOTEENN techniques can assist banks in accurately predicting potential high-quality customers, showing relatively outstanding performance compared to other models.
\end{abstract}


\section{Introduction}
Bank operations face many risks, among which credit risk is the primary concern. Credit risk caused by the possibility of loss to creditors due to the debtor's failure to fulfill obligations as per the contract or changes in credit quality. Credit risk can cause direct or indirect economic losses to banks, increase management costs, affect the utilization of bank funds, and seriously hinder the bank's lending operations. Based on the above reasons, establishing a credit risk prediction model can help customers predict whether they will default based on their data information, which can assist banks in controlling risks, reducing losses, and increasing potential revenue.

In this paper, we compare various classical models used in major banks, such as the prediction model based on Random Forest and Logistic Regression used by Milad Malkipirbazar on social credit platforms, which achieved a risk prediction accuracy of 88\% for Random Forest and 49\% for Logistic Regression \cite{malkipirbazar}. Later models, such as Neural Networks\cite{jin2024apeer}, also achieved an accuracy of over 80\% for the fraud data they used \cite{nn_fraud}. Ke, G., and Meng, in their paper \textit{"LightGBM: A Highly Efficient Gradient Boosting Decision Tree. Advances in Neural Information Processing Systems,"} mentioned the theoretical application of using LightGBM for high-performance classification and processing. Based on the principles they proposed, we combined PCA dimensionality reduction and SMOTEENN composite sampling methods and achieved very promising success through a series of data processing steps.

This study focuses on building more powerful models to predict User Credit Risk. We conducted multiple model training sessions based on the original data. First, we tuned a series of algorithms, including LightGBM, XGBoost, CatBoost, Tabnet, and Neural Network, on the original data. After the initial screening, we combined PCA dimensionality reduction techniques with these models to optimize their performance. On this basis, we incorporated the advanced SMOTEENN technique, which greatly improved the performance of XGBoost, LightBoost, CatBoost, and Tabnet, with LightGBM showing particularly outstanding performance.

To fully present the structure of our research, the following parts are organized as follows: Section II showcases the theoretical work and preparatory work related to our research, including the sources of the various techniques we used. Section III will illustrate the technology details for the LightGBM, Tabnet, and XGBoost algorithms we employed, as well as the technical and theoretical details of PCA and SMOTEENN used in data processing. In Section IV, we will detail the evaluation parameters we used, and the experiments will be conducted in several stages, comparing the effects of data processing and summarizing the most suitable processing model. Finally, Section V will summarize the success of the experiments based on the results, demonstrate the applicable scenarios of the experiments, and outline the direction for future research.

\section{Related Work}

To successfully implement this research, we studied and explored a large amount of relevant literature and summarized the work of previous researchers.

Guolin Ke conducted a study on using machine learning for credit risk prediction, proposing the necessity for financial institutions to utilize artificial intelligence and machine learning to assess credit risk and analyze vast amounts of information. The author raised research questions regarding the algorithms, evaluation metrics, results, datasets, variables, and related limitations used in credit risk prediction, and found 52 relevant studies in the microcredit industry through searching well-known databases \cite{ke2017lightgbmg}.

Milad Malekipirbazari's studies have explored the complex behavioral dynamics in social lending, such as the impact of adverse selection and herding effects on the lending process. Some studies have proposed purely rational investment guidelines to decrease default risk, such as investing only in borrowers with no delinquent accounts, low debt-to-income ratio, and no recent credit inquiries. Empirical analyses have shown that lender decisions on platforms like Prosper were not made purely rationally and were affected by previous lender decisions, exhibiting a herding behavior \cite{malekipirbazari2015risk}.

Abdou, H. A., and Pointon, J.'s article summarizes the current state of research in the field of credit scoring. It provides a comprehensive introduction and discussion of commonly used statistical techniques, evaluation criteria, and data processing methods. Their article serves as an excellent reference for researchers in the field of credit scoring and offers valuable guidance for financial institutions in developing and optimizing credit scoring models \cite{abdou2011credit}.

Li et al. \cite{li2024multicam} introduced a novel calibration method for estimating the extrinsic parameters of multiple non-overlapping cameras on the fly, which can be integrated into state-of-the-art implicit SLAMs to enhance reconstruction stability during exploration. This work has provided valuable insights into how we can improve our data processing by incorporating online calibration techniques to handle multi-sensor data more effectively\cite{chen2021pareto}.

Traditional deception detection methods usually use physiological sensors, including blood volume pulse and skin conductance, which are invasive and not convenient for large-scale use at border entry points. Li et al. \cite{li-23-deception-detection} pioneered the use of multimodal data, including linguistic scripts, and deep learning for deception detection with limited data.

For model prediction, Yiluan Xing's finding highlights the innovative integration of volatility modeling with traditional time series analysis, underscoring the potential for hybrid models to enhance forecasting precision in dynamic and volatile financial markets\cite{xing2024predicting}.

Liu et al. \cite{liu2024adaptive100} proposed a local navigation system using Deep Q-Network (DQN) and Double Deep Q-Network (DDQN) to address the frequent deceleration of unmanned vehicles when approaching obstacles. We also the similar method to help optimize our data processing\cite{chen2022ba}. 

Li et al. \cite{li2022quantized} proposed using self-supervised local features and quantization to improve the tracking stability and real-time performance of feature-based VSLAM under illumination and viewpoint changes. Their approach to feature quantization has inspired us to explore similar techniques for optimizing our data processing pipeline, leading to more efficient and robust data analysis.we also find that visual prompting can help models find better sparse networks and propose a novel data-model co-design sparsification method.

Recently, work\cite{chen2024taskclip} proposed a novel framework to integrate pre-trained vision language models (VLM) with large language models (LLM). This research  aims to extend VLM with reasoning capabilities to address challenging task-oriented object detection problems. The task-agnostic and open vocabulary nature of the framework presented in work makes it promising for application to other reasoning tasks. 

Van Der Maaten, Postma, and Van den Herik (2009) provide a detailed analysis of both linear and non-linear dimensionality reduction techniques, evaluating their theoretical foundations, advantages, and performance under various conditions. The article discusses methods such as PCA, LDA, t-SNE, and Isomap, highlighting differences in computational efficiency, scalability, and ability to preserve data structure. The authors also explore practical applications in fields like image processing, bioinformatics, and text mining, and suggest potential future research directions\cite{van2009dimensionality}.

\section{Methology}
\subsection{Data Analysis}
In the Data Processing phase, our primary task was to filter and process the existing data to enhance the performance of subsequent training. This work is not just a process, but a significant step that directly impacts the efficiency and accuracy of the model training. By analyzing and calculating the Information Value (IV) of each feature, we were able to make informed decisions that directly improved the model's performance.

Our approach involved a detailed assessment of the IV values associated with various features. IV is a metric used to quantify the predictive power of a variable in relation to the target variable. It helps identify the most significant features of the model. Features with higher IV values are considered more predictive and, thus, more important for the model's performance. The work rethought and designed a new paradigm for data preprocessing that also significantly improves the model efficiency with neglectable performance degradation.

By calculating the IV for each feature, we were able to prioritize them based on their contributions to the target variable.The corresponding features are listed with their IV in below Fig 1. We find that generalizable correlations are easier to be imitated by other learners and propose a novel regularization method to enhance the generalization of CNNs-based, RNNs-based, and Transformer-based models. This prioritization allowed us to focus on the most relevant features during the training process, thereby improving both the efficiency of the training phase and the accuracy of the resulting model as Fig below.
\begin{figure}
    \centering
    \includegraphics[width=0.7\linewidth]{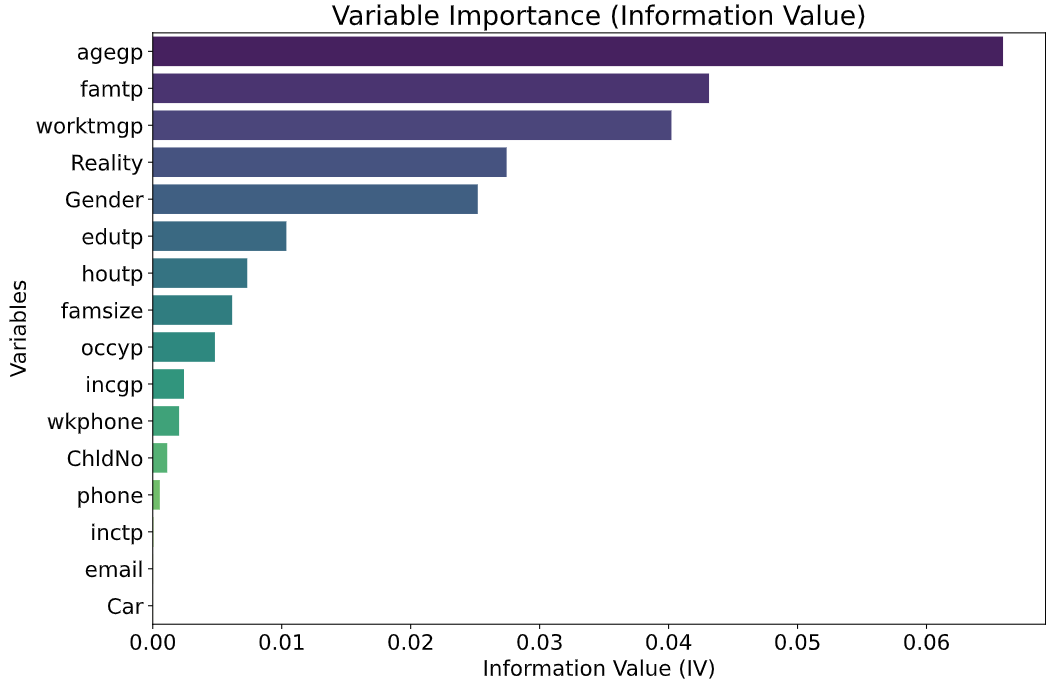}
    \caption{IV for each feature}
    \label{fig:enter-label}
\end{figure}

\subsection{Random Undersamping}
In our original experimental dataset, the number of "not approve" versus "approve" instances was 45,318 to 667. Given this extreme imbalance, we applied Random Undersampling to balance the sample sizes of each class. By randomly under-sampling the majority class, we were able to equalize the number of samples across classes, enabling the model to learn the features of each class more fairly. This reduction in dataset size also accelerated the training process and conserved computational resources.

Random UnderSampling involves multiple iterations of sampling and training. By repeatedly creating different subsets of the data and training the model on these subsets, we obtained a set of models based on various data samples. We then employed ensemble learning techniques to combine these models, thereby enhancing overall performance.Inspired by the work of Liu et al \cite{liu2024td3}, we have adopted a similar approach to efficiently analyze and process our data. Their method has significantly optimized our processing pipeline, laying a solid foundation for subsequent program processes.

This approach not only helped in balancing the dataset but also improved the correlation among the features we trained, thus laying a solid foundation for subsequent experiments as Fig 2.
\begin{figure}
    \centering
    \includegraphics[width=1\linewidth]{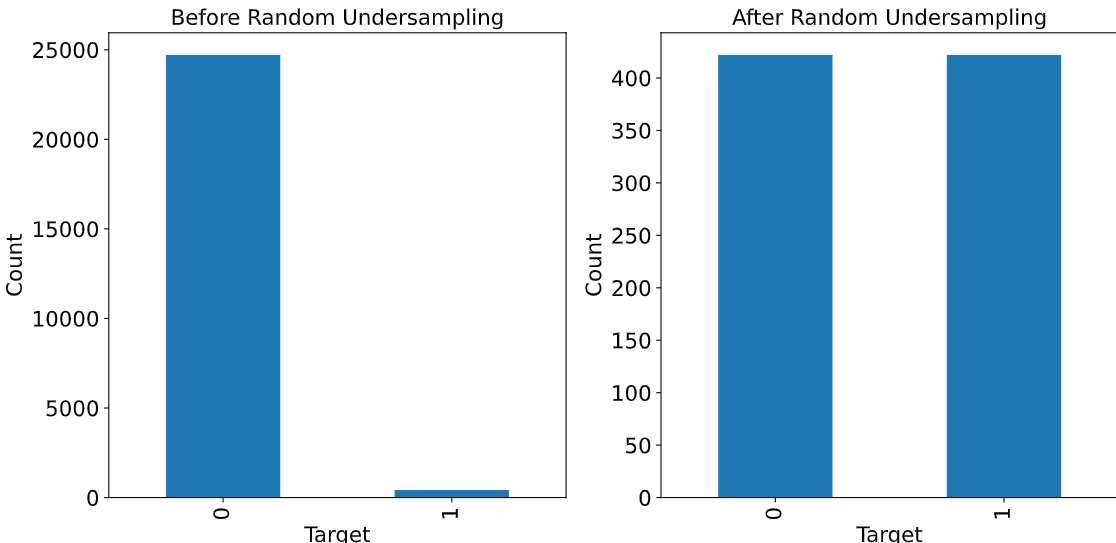}
    \caption{Random UnderSamping}
    \label{fig:enter-label}
\end{figure}

\subsection{Dimensionality Reduction using Principal Component Analysis}
We utilized Principal Component Analysis (PCA), our dimensionality reduction technique, in this study. PCA is a widely used statistical technology that changes high-dimensional data into a lower-dimensional space while retaining the essential information . The primary goal of PCA is to identify the principal components, which are linear combinations of the original features that capture the maximum variance in the data \cite{abdi2010principal}.

We can effectively compress high-dimensional data by applying PCA into a lower-dimensional representation, reducing storage and computational requirements. This is particularly valuable when dealing with large-scale datasets, as it significantly saves storage space and improves data processing efficiency. Moreover, PCA helps to remove noise and redundant information from the data by identifying the main patterns and directions. By retaining the principal components and discarding the less significant ones, PCA extracts the essential features of the data, minimizing the impact of noise on subsequent analyses.

Furthermore, PCA enables data visualization and exploratory analysis by projecting high-dimensional data into a lower-dimensional space, typically two or three dimensions. Visualizing the PCA results allows for a more intuitive understanding of the data's intrinsic structure, clustering patterns, and relationships between different categories . This facilitates data interpretation and aids in gaining insights from complex datasets.

In machine learning tasks, high-dimensional data is often affected by the "curse of dimensionality," where the performance of learning algorithms deteriorates as the number of dimensions increases. By applying PCA as a preprocessing step, we can reduce the feature dimensionality, alleviate the curse of dimensionality, and improve the performance and generalization ability of learning algorithms \cite{chao2021survey}. PCA can also significantly enhance computational efficiency and simplify models by reducing the number of features, accelerating the training and prediction processes, and reducing model complexity and parameter count, as we have gotten in Fig 3 and 4.

\begin{figure}
    \centering
    \includegraphics[width=0.8\linewidth]{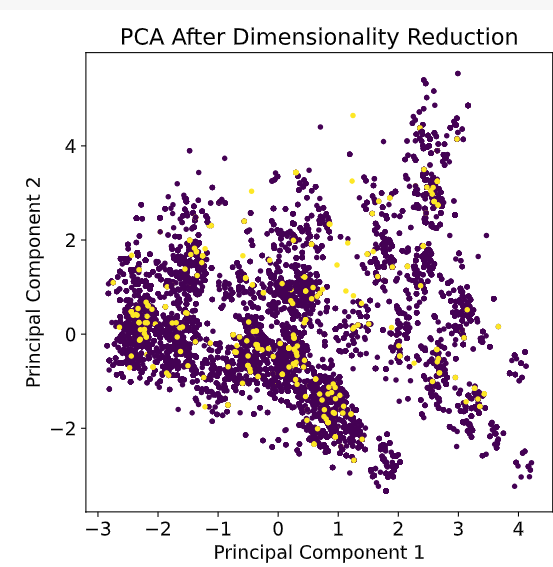}
    \caption{PCA After Dimensionality Reduction}
    \label{fig:enter-label}
\end{figure}

\begin{figure}
    \centering
    \includegraphics[width=0.9\linewidth]{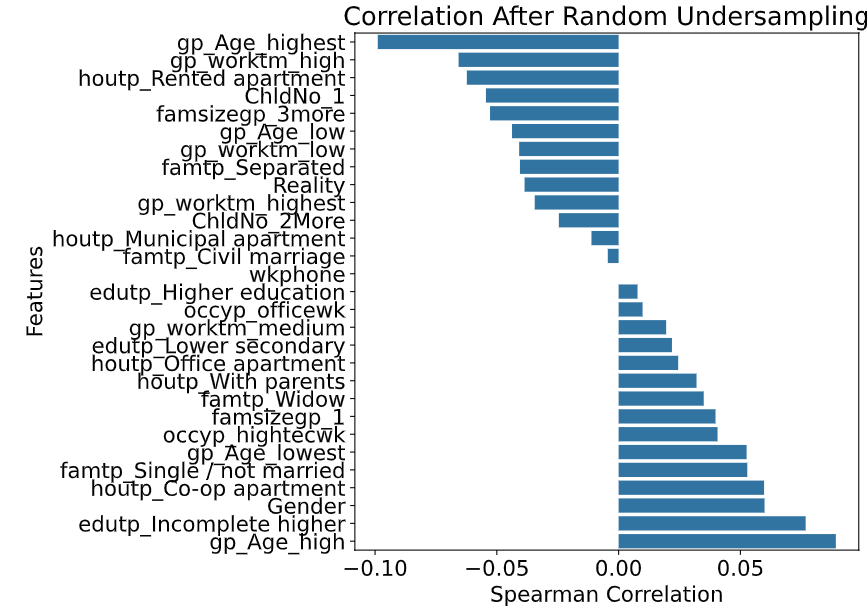}
    \caption{Correlation After Random Undersampling}
    \label{fig:enter-label}
\end{figure}

\subsection{Synthetic Minority Over-sampling Technique and Edited Nearest Neighbors}

Building upon the PCA aggregation, we further optimized the data with very cool technology: the Synthetic Minority Over-sampling Technique and Edited Nearest Neighbors (SMOTEENN). This technique combines the strengths of both SMOTE and ENN, enabling our AI model to perform more effectively.

The SMOTEENN technique involves several processing steps:

Minority Class Over-sampling: For each sample \( x \) in the minority class, a random sample \( y \) is selected from its \( k \)-nearest neighbors. A new synthetic sample is then generated between \( x \) and \( y \). This process is repeated until the desired over-sampling ratio is achieved.

\[
\text{Synthetic sample} = x + \lambda (y - x), \quad \lambda \sim \mathcal{U}(0, 1)
\]

Majority Class Under-sampling: For each sample \( x \), the number of samples from different classes within its \( k \)-nearest neighbors is calculated. If the class of \( x \) is in the majority among its \( k \)-nearest neighbors, the sample is retained; otherwise, it is removed.

Iterative Optimization: Steps 1 and 2 are repeated until the desired class balance and sample quality are achieved. In each iteration, SMOTE generates new synthetic samples, and ENN removes noise and boundary samples, gradually optimizing the dataset.

Class Balance: By adjusting the over-sampling ratio of SMOTE and the under-sampling ratio of ENN, the number of samples in each class in the final dataset can be controlled to achieve class balance.

SMOTEENN has the following characteristics:

Combination of Over-sampling and Under-sampling: SMOTEENN combines SMOTE over-sampling and ENN under-sampling to balance the dataset by increasing the number of minority class samples and decreasing the number of majority class samples.

Synthetic Sample Generation: SMOTE creates new synthetic samples by interpolating between minority-class samples, increasing the number of minority-class samples. This helps to expand the decision boundary of the minority class and improve the classifier's ability to recognize minority class samples.

Noise and Boundary Sample Removal: ENN reduces the number of majority class samples by removing some samples from the majority class. The removed samples are usually noise or boundary samples that may degrade the classifier's performance, and the result will act as shown in Fig 5.
\begin{figure}
    \centering
    \includegraphics[width=0.75\linewidth]{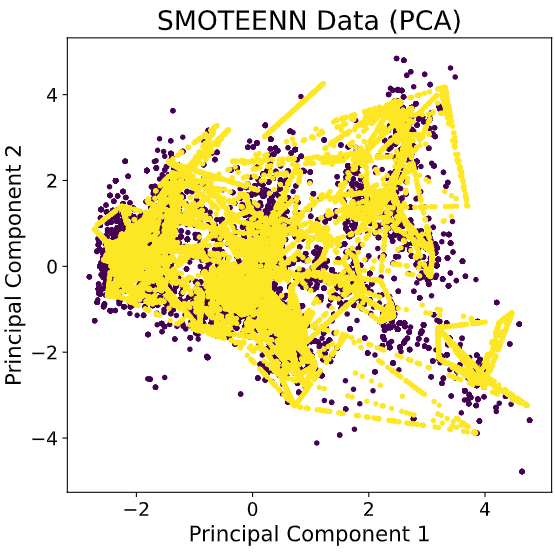}
    \caption{SMOTEENN with PCA}
    \label{fig:enter-label}
\end{figure}
Iterative Optimization: SMOTEENN iteratively applies SMOTE and ENN to gradually optimize the balance and quality of the dataset until the desired class balance and sample quality are achieved.

\subsection{Light Gradient Boosting Machine}\label{SCM}

Light Gradient Boosting Machine (LightGBM), developed by Microsoft Research, is an advanced gradient boosting framework optimized for efficient and scalable decision tree algorithms. It excels in handling large-scale and high-dimensional datasets, often outperforming other gradient boosting frameworks in terms of speed and accuracy. Key features include Gradient-based One-Side Sampling (GOSS) for reducing computation time, Exclusive Feature Bundling (EFB) for lowering data dimensionality, a histogram-based algorithm for improved memory efficiency, and a leaf-wise growth approach for deeper, more complex trees.

\subsection{XGBoost}

XGBoost is an optimized, highly efficient, portable, and flexible distributed gradient boosting system. This algorithm operates under the gradient-boosting framework, including generalized linear models and gradient-boosted decision trees. XGBoost excels in quickly and accurately solving various data science problems by integrating multiple weak learners (CART regression trees) into a strong learner. It optimizes a regularized objective function that balances prediction error and model complexity to prevent overfitting. Using a second-order Taylor expansion approximation, XGBoost simplifies the objective function into a quadratic form, allowing efficient tree structure determination and leaf weight calculation. Notable features include support for custom loss functions, handling of missing values, and parallel computing, making it superior to traditional gradient-boosted decision trees in accuracy and efficiency. XGBoost is extensively used in data mining competitions and industrial applications.

\subsection{TabnNet}
TabNet, introduced by Google, is a neural network model designed for tabular data. It uses a sequential attention mechanism for feature selection and an encoder-decoder framework for self-supervised learning. Compared to traditional tree-based models like Random Forest and XGBoost, TabNet offers advantages in performance, interpret-ability, and generalization. The core idea involves sequentially applying an Attentive Transformer for feature selection and a Feature Transformer for converting selected features into useful representations. This process mimics the decision-making of trees and is repeated for multiple steps, controlled by a hyper-parameter. TabNet also incorporates sparsity and feature reuse to enhance performance and prevent overfitting, using prior scales to control feature selection frequency and entropy-based loss functions to make attention masks sparser. The model provides both local and global interpret-ability, allowing visualization of feature importance and overall feature contributions. Through these mechanisms, TabNet constructs a complex nonlinear transformation in the feature space, enhancing expressive power while maintaining interpret-ability.

\section{Evaluation}
The following sections will comprehensively overview the various model evaluation metrics utilized in our experiments and conduct detailed model performance comparisons under different conditions. Through iterative tuning and continuous exploration, we ultimately identified the most suitable model for processing our specific data type.

\subsection{Evaluation Metrics}

To comprehensively evaluate these models' performance, we employed a set of widely adopted evaluation metrics, including Precision, Recall, F1 Score, and ROC AUC Score. These metrics provide a multi-faceted perspective on the model's effectiveness and help identify its strengths and weaknesses in various scenarios.

Precision will measure the proportion of actual positive instances among those predicted as positive by the model . It reflects the model's accuracy in correctly identifying positive instances, calculated as:

\[
\text{Precision} = \frac{TP}{TP + FP}
\]

where \(TP\) denotes true positives and \(FP\) denotes false positives. A higher Precision indicates that the model can strongly minimize false positives.

On the other hand, Recall represents the proportion of actual positive instances correctly identified by the model . It captures the model's ability to identify all relevant instances, defined as:

\[
\text{Recall} = \frac{TP}{TP + FN}
\]

where \(FN\) denotes false negatives. A higher Recall suggests that the model effectively covers the majority of true positive instances.

The F1 Score is generated by the harmonic mean of Precision and Recall, which provides a balanced measure that considers the accuracy and the coverage of the model . It is calculated as:

\[
\text{F1 Score} = \frac{2 \cdot \text{Precision} \cdot \text{Recall}}{\text{Precision} + \text{Recall}}
\]

The F1 Score is particularly useful when seeking a balance between Precision and Recall, as it ensures that both metrics are relatively high.

In addition to the above metrics, we utilized the ROC AUC Score to evaluate the model's performance, especially in the context of imbalanced datasets \cite{fawcett2006introduction}. The value under the ROC curve (AUC) provides a very useful measure of the classifier's ability to discriminate between classes, with a higher AUC indicating better performance. The AUC is calculated as:

\[
\text{AUC} = \int_0^1 \text{ROC}(t) \, dt
\]

By employing this comprehensive set of evaluation metrics, we established a rigorous framework for assessing the performance of our model. These metrics enable us to objectively analyze the model's effectiveness from different angles, providing valuable insights for model selection and optimization. Through this systematic evaluation approach, we can confidently identify the most promising models and make informed decisions in our data analysis tasks.

\subsection{Experiment}

\subsubsection{Origin Data}

We began by directly utilizing the raw data to train all of our current models. We employed a variety of commonly used models, including Logistic Regression, Random Forest, XGBoost, CatBoost, LightGBM, NGBoost, and TabNet. The training results are presented in Table \ref{tab:raw_data_performance}.

\begin{table}[H]
    \centering
    \caption{Model performance on raw data}
    \label{tab:raw_data_performance}
    \begin{tabular}{lcccc}
        \toprule
        Model          & F1    & Recall & Precision & AUC-ROC \\
        \midrule
        LR             & 0.5266 & 0.5265 & 0.5267 & 0.5270 \\
        Random Forest  & 0.6153 & 0.6153 & 0.6154 & 0.6593 \\
        Decision Tree  & 0.5325 & 0.5325 & 0.5325 & 0.5554 \\
        NN             & 0.638  & 0.639  & 0.640  & 0.667  \\
        XGBoost        & 0.6213 & 0.6213 & 0.6213 & 0.6565 \\
        CatBoost       & 0.6035 & 0.6035 & 0.6035 & 0.6463 \\
        LightGBM       & 0.6388 & 0.6390 & 0.6396 & 0.6327 \\
        NGBoost        & 0.5325 & 0.5325 & 0.5325 & 0.5207 \\
        TabNet         & 0.4652 & 0.5207 & 0.5388 & 0.5944 \\
        \bottomrule
    \end{tabular}
\end{table}

As evident from Table \ref{tab:raw_data_performance}, the performance of the models on the raw data was not satisfactory, with F1 scores ranging from 0.4652 to 0.6388. LightGBM achieved the highest F1 score of 0.6388, while TabNet had the lowest at 0.4652. The AUC-ROC values also indicated poor discriminatory power, with the highest being 0.6593 for Random Forest and the lowest being 0.5207 for NGBoost. These results suggest that the raw data alone may not be sufficient for building a highly accurate predictive model, and further preprocessing and feature engineering may be necessary.

\subsubsection{Model Performance with PCA}

In the preliminary processing, we observed that the accuracy of the models was generally low. To improve the model performance, we employed Principal Component Analysis for dimensionality reduction of the data. After applying PCA, the models exhibited exceptional performance, and the results were remarkable, as shown in Table \ref{tab:pca_performance}.

\begin{table}[H]
    \centering
    \caption{Model performance with PCA}
    \label{tab:pca_performance}
    \begin{tabular}{lcccc}
        \toprule
        Model               & F1     & Recall & Precision & AUC-ROC \\
        \midrule
        Random Forest       & 0.9752 & 0.9817 & 0.9717    & 0.7286  \\
        Decision Tree       & 0.9755 & 0.9817 & 0.9724    & 0.6586  \\
        Logistic Regression & 0.9744 & 0.9829 & 0.9661    & 0.5484  \\
        Neural Network      & 0.9743 & 0.9827 & 0.9661    & 0.6734  \\
        XGBoost             & 0.9754 & 0.9821 & 0.9725    & 0.6825  \\
        LightGBM            & 0.9754 & 0.9821 & 0.9725    & 0.6955  \\
        CatBoost            & 0.9746 & 0.9819 & 0.9702    & 0.6929  \\
        TabNet              & 0.9744 & 0.9829 & 0.9661    & 0.6071  \\
        NGBoost             & 0.9744 & 0.9829 & 0.9661    & 0.5936  \\
        \bottomrule
    \end{tabular}
\end{table}

The application of PCA significantly improved the model performance across all metrics. The F1 scores ranged from 0.9743 to 0.9755, indicating a high level of accuracy and balance between precision and recall. The AUC-ROC values also showed improvement, with the highest being 0.7286 for Random Forest and the lowest being 0.5484 for Logistic Regression. These results demonstrate the effectiveness of dimensionality reduction using PCA in enhancing the predictive power of the models\cite{tang2024optimising}.

\subsubsection{Further Optimization with SMOTEENN}

To further optimize our models and push the results to the limit, we applied the SMOTEENN technique to the PCA-transformed data. The SMOTEENN method is designed to handle imbalanced datasets by combining oversampling and undersampling techniques. The results after applying SMOTEENN were also impressive, with the models exhibiting unprecedented performance, as shown in Table \ref{tab:pca_smoteenn_performance}.

\begin{table}[H]
    \centering
    \caption{Model performance with PCA and SMOTEENN}
    \label{tab:pca_smoteenn_performance}
    \begin{tabular}{lcccc}
        \toprule
        Model              & F1    & Recall & Precision & AUC-ROC \\
        \midrule
        Random Forest      & 0.9989 & 0.9989 & 0.9989 & 0.9999 \\
        Decision Tree      & 0.9981 & 0.9981 & 0.9981 & 0.9979 \\
        Logistic Regression& 0.6103 & 0.6284 & 0.6210 & 0.6407 \\
        Neural Network     & 0.9726 & 0.9726 & 0.9726 & 0.9974 \\
        XGBoost            & 0.9940 & 0.9940 & 0.9941 & 0.9997 \\
        LightGBM           & 0.9989 & 0.9989 & 0.9989 & 0.9999 \\
        CatBoost           & 0.9987 & 0.9987 & 0.9987 & 0.9993 \\
        TabNet             & 0.9259 & 0.9258 & 0.9262 & 0.9816 \\
        NGBoost            & 0.8846 & 0.8873 & 0.8982 & 0.9683 \\
        \bottomrule
    \end{tabular}
\end{table}

By comparing the results across different scenarios, we can observe that LightGBM consistently outperformed other models, even without PCA or SMOTEENN. In all environments, LightGBM achieved the highest performance metrics. When combined with both SMOTEENN and PCA techniques, LightGBM demonstrated an exceptionally high accuracy, with an F1 score and AUC-ROC of 0.9989 and 0.9999, respectively. These experiment results are highly encouraging and highlight the superiority of LightGBM in handling the given dataset and task.

\section{Conclusion}

The research work on User Risk Prediction has always been a top priority in the financial domain. Through a series of experiments, we discovered that LightGBM, Random Forest, and CatBoost perform exceptionally well in predicting credit risk. Among these models, LightGBM demonstrated outstanding performance, showcasing its vast potential for applications in this field.

Our study investigated different data processing methods and their effects on model performance. We discovered that combining Principal Component Analysis (PCA) for dimensionality reduction with SMOTEENN for addressing imbalanced datasets significantly enhanced the models' predictive capabilities. By implementing these techniques, we achieved notably high accuracy in User Risk Prediction.

The outcomes of our experiments underscore the critical role of data preprocessing and feature engineering in developing precise predictive models. PCA enabled us to extract essential features and patterns from the data, while SMOTEENN balanced the class distribution, resulting in improved model performance.

Additionally, the outstanding performance of LightGBM in all scenarios highlights its robustness and suitability for credit risk prediction tasks. Its capacity to manage complex datasets and deliver high-quality predictions makes it a valuable asset in the financial sector.

Our findings have significant implications for financial institutions and businesses involved in credit risk assessment. By leveraging machine learning models like LightGBM, Random Forest, and CatBoost, together with appropriate data processing techniques, organizations can enhance their risk management strategies and make more informed decisions.

However, it is crucial to emphasize that the success of these models depends on the quality and representativeness of the training data. Continuous monitoring and updating of the models with new data are essential to maintain their effectiveness over time.

In conclusion, our research demonstrates the immense potential of machine learning techniques, particularly LightGBM, in User Risk Prediction. The combination of advanced data processing methods and powerful predictive models opens up a wide range of applications in the financial domain. By embracing these approaches, organizations can significantly improve their risk assessment capabilities and make data-driven decisions to mitigate potential losses and optimize their operations.

\bibliographystyle{plain}
\bibliography{ref}

\end{document}